# One Sense per Collocation and Genre/Topic Variations

David Martinez
IXA NLP Group
University of the Basque Country
649 pk. 20.080
Donostia. Spain
jibmaird@si.ehu.es

Eneko Agirre
IXA NLP Group
University of the Basque Country
649 pk. 20.080
Donostia. Spain
eneko@si.ehu.es

## Abstract

This paper revisits the one sense per collocation hypothesis using fine-grained sense distinctions and two different corpora. We show that the hypothesis is weaker for fine-grained sense distinctions (70% vs. 99% reported earlier on 2-way ambiguities). We also show that one sense per collocation does hold across corpora, but that collocations vary from one corpus to the other, following genre and topic variations. This explains the low results when performing word sense disambiguation across corpora. In fact, we demonstrate that when two independent corpora share a related genre/topic, the word sense disambiguation results would be better. Future work on word sense disambiguation will have to take into account genre and topic as important parameters on their models.

## Introduction

In the early nineties two famous papers claimed that the behavior of word senses in texts adhered to two principles: one sense per discourse (Gale et al., 1992) and one sense per collocation (Yarowsky, 1993).

These hypotheses were shown to hold for some particular corpora (totaling 380 Mwords) on words with 2-way ambiguity. The word sense distinctions came from different sources (translations into French, homophones, homographs, pseudo-words, etc.), but no dictionary or lexical resource was linked to them. In the case of the one sense per collocation paper, several corpora were used, but nothing is said on whether the collocations hold across corpora.

Since the papers were published, word sense disambiguation has moved to deal with fine-grained sense distinctions from widely recognized semantic lexical resources; ontologies like Sensus, Cyc, EDR, WordNet, EuroWordNet, etc. or machine-readable dictionaries like OALDC, Webster's, LDOCE, etc. This is due, in part, to the availability of public hand-tagged material, e.g. SemCor (Miller et al., 1993) and the DSO collection (Ng & Lee, 1996). We think that the old hypotheses should be tested under the conditions of this newly available data. This paper focuses on the DSO collection, which was tagged with WordNet senses (Miller et al. 1990) and comprises sentences extracted from two different corpora: the balanced Brown Corpus and the Wall Street Journal corpus.

Krovetz (1998) has shown that the one sense per discourse hypothesis does not hold for fine-grained senses in SemCor and DSO. His results have been confirmed in our own experiments. We will therefore concentrate on the one sense per collocation hypothesis, considering these two questions:

- Does the collocation hypothesis hold across corpora, that is, across genre and topic variations (compared to a single corpus, probably with little genre and topic variations)?
- Does the collocation hypothesis hold for fine-grained sense distinctions (compared to homograph level granularity)?

The experimental tools to test the hypothesis will be decision lists based on various kinds of collocational information. We will compare the performance across several corpora (the Brown Corpus and Wall Street Journal parts of the DSO collection), and also across different sections of the Brown Corpus, selected according to the genre and topics covered. We will also perform a direct comparison, using agreement statistics, of the collocations used and of the results obtained.

This study has special significance at this point of word sense disambiguation research. A recent study (Agirre & Martinez, 2000) concludes that, for currently available hand-tagged data, the precision is limited to around 70% when tagging all words in a running text. In the course of extending available data, the efforts to use corpora tagged by independent teams of researchers have been shown to fail (Ng et al., 1999), as have failed some tuning experiments (Escudero et al., 2000), and an attempt to use examples automatically acquired from the Internet (Agirre & Martinez, 2000). All these studies obviated the fact that the examples come from different genre and topics. Future work that takes into account the conclusions drawn in this paper will perhaps be able to automatically extend the number of examples available and tackle the acquisition problem.

The paper is organized as follows. The resources used and the experimental settings are presented first. Section 3 presents the collocations considered and Section 4 explains how decision lists have been adapted to n-way ambiguities. Sections 5 and 6 show the in-corpus and cross-corpora experiments, respectively. Section 7 discusses the effect of drawing training and testing data from the same documents. Section 8 evaluates the impact of genre and topic variations, which is further discussed in Section 9. Finally, Section 10 presents some conclusions.

# 1 Resources used

The **DSO** collection (Ng and Lee, 1996) focuses on 191 frequent and polysemous words (nouns and verbs), and contains around 1,000 sentences per word. Overall, there are 112,800 sentences, where 192,874 occurrences of the target words were hand-tagged with WordNet senses (Miller et al., 1990).

The DSO collection was built with examples from the **Wall Street Journal** (WSJ) and **Brown Corpus** (BC). The Brown Corpus is balanced, and the texts are classified according some predefined categories (cf. Table 1). The examples from the Brown Corpus comprise 78,080 occurrences of word senses, and the examples from the WSJ 114,794 occurrences.

The sentences in the DSO collection were tagged with parts of speech using TnT (Brants, 2000) trained on the Brown Corpus itself.

| |
|---|
| A. Press: Reportage |
| B. Press: Editorial |
| C. Press: Reviews (theatre, books, music, dance) |
| D. Religion |
| E. Skills and Hobbies |
| F. Popular Lore |
| G. Belles Lettres, Biography, Memoirs, etc. |
| H. Miscellaneous |
| J. Learned |
| K. General Fiction |
| L. Mystery and Detective Fiction |
| M. Science Fiction |
| N. Adventure and Western Fiction |
| P. Romance and Love Story |
| R. Humor |

**Table 1:** List of categories of texts from the Brown Corpus, divided into informative prose (top) and imaginative prose (bottom).

## 1.1 Categories in the Brown Corpus and genre/topic variation

The Brown Corpus manual (Francis & Kucera, 1964) does not detail the criteria followed to set the categories in Table 1:

*The samples represent a wide range of styles and varieties of prose... The list of main categories and their subdivisions was drawn up at a conference held at Brown University in February 1963.*

These categories have been previously used in genre detection experiments (Karlgren & Cutting, 1994), where each category was used as a genre. We think that the categories not only reflect genre variations but also topic variations (e.g. the *Religion* category follows topic distinctions rather than genre). Nevertheless we are aware that some topics can be covered in more than one category. Unfortunately there are no topically tagged corpus which also have word sense tags. We thus speak of genre and topic variation, knowing that further analysis would be needed to measure the effect of each of them.

## 2 Experimental setting

In order to analyze and compare the behavior of several kinds of collocations (cf. Section 3), Yarowsky (1993) used a measure of entropy as well as the results obtained when tagging held-out data with the collocations organized as decision lists (cf. Section 4). As Yarowsky shows, both measures correlate closely, so we only used the experimental results of decision lists.

When comparing the performance on decision lists trained on two different corpora (or sub-corpora) we always take an equal amount of examples per word from each corpora. This is done to discard the amount-of-data factor.

As usual, we use 10-fold cross-validation when training and testing on the same corpus. No significance tests could be found for our comparison, as training and test sets differ.

Because of the large amount of experiments involved, we focused on 21 verbs and nouns (cf. Table 2), selected from previous works (Agirre & Martinez, 2000; Escudero et al., 2000).

## 3 Collocations considered

For the sake of this work we take a broad definition of collocations, which were classified in three subsets: local content word collocations, local part-of-speech and function-word collocations, and global content-word collocations. If a more strict linguistic perspective was taken, rather than collocations we should speak about co-occurrence relations. In fact, only local content word collocations would adhere to this narrower view.

We only considered those collocations that could be easily extracted form a part of speech tagged corpus, like word to left, word to right, etc. Local content word collocations comprise bigrams (word to left, word to right) and trigrams (two words to left, two words to right and both words to right and left). At least one of those words needs to be a content word. Local function-word collocations comprise also all kinds of bigrams and trigrams, as before, but the words need to be function words. Local PoS collocations take the Part of Speech of the words in the bigrams and trigrams. Finally global content word collocations comprise the content words around the target word in two different contexts: a window of 4 words around the target word, and all the words in the sentence. Table 3 summarizes the collocations used. These collocations have been used in other word sense disambiguation research and are also referred to as *features* (Gale et al., 1993; Ng & Lee, 1996; Escudero et al., 2000).

Compared to Yarowsky (1993), who also took into account grammatical relations, we only share the *content-word-to-left* and the *content-word-to-right* collocations.

| Word | PoS | #Senses | #Ex. BC | #Ex. WSJ |
|---|---|---|---|---|
| Age | N | 5 | 243 | 248 |
| Art | N | 4 | 200 | 194 |
| Body | N | 9 | 296 | 110 |
| Car | N | 5 | 357 | 1093 |
| Child | N | 6 | 577 | 484 |
| Cost | N | 3 | 317 | 1143 |
| Head | N | 28 | 432 | 434 |
| Interest | N | 8 | 364 | 1115 |
| Line | N | 28 | 453 | 880 |
| Point | N | 20 | 442 | 249 |
| State | N | 6 | 757 | 706 |
| Thing | N | 11 | 621 | 805 |
| Work | N | 6 | 596 | 825 |
| Become | V | 4 | 763 | 736 |
| Fall | V | 17 | 221 | 1227 |
| Grow | V | 8 | 243 | 731 |
| Lose | V | 10 | 245 | 935 |
| Set | V | 20 | 925 | 355 |
| Speak | V | 5 | 210 | 307 |
| Strike | V | 17 | 159 | 95 |
| Tell | V | 8 | 740 | 744 |

**Table 2:** Data for selected words. Part of speech, number of senses and number of examples in BC and WSJ are shown.

| Local content word collocations | | |
|---|---|---|
| Word-to-left | | Content Word |
| Word-to-right | | Content Word |
| Two-words-to-left | | At least one Content Word |
| Two-words-to-right | | |
| Word-to-right-and-left | | |
| Local PoS and function word collocations | | |
| Word-to-left | PoS | Function Word |
| Word-to-right | PoS | Function Word |
| Two-words-to-left | PoS | Both Function Words |
| Two-words-to-right | PoS | |
| Word-to-right-and-left | PoS | |
| Global content word collocations | | |
| Word in Window of 4 | | Content Word |
| Word in sentence | | |

**Table 3:** Kinds of collocations considered

We did not lemmatize content words, and we therefore do take into account the form of the target word. For instance, *governing body* and *governing bodies* are different collocations for the sake of this paper.

## 4 Adaptation of decision lists to n-way ambiguities

Decision lists as defined in (Yarowsky, 1993; 1994) are simple means to solve ambiguity problems. They have been successfully applied to accent restoration, word sense disambiguation and homograph disambiguation (Yarowsky, 1994; 1995; 1996). In order to build decision lists the training examples are processed to extract the features (each feature corresponds to a kind of collocation), which are weighted with a log-likelihood measure. The list of all features ordered by log-likelihood values constitutes the decision list. We adapted the original formula in order to accommodate ambiguities higher than two:

$$weight(sense_i, feature_k) = Log(\frac{\Pr(sense_i \mid feature_k)}{\sum_{j \neq i} \Pr(sense_j \mid feature_k)})$$

When testing, the decision list is checked in order and the feature with highest weight that is present in the test sentence selects the winning word sense. For this work we also considered negative weights, which were not possible on two-way ambiguities.

The probabilities have been estimated using the maximum likelihood estimate, smoothed using a simple method: when the denominator in the formula is 0 we replace it with 0.1. It is not clear how the smoothing technique proposed in (Yarowsky, 1993) could be extended to n-way ambiguities.

More details of the implementation can be found in (Agirre & Martinez, 2000).

## 5 In-corpus experiments: collocations are weak (80%)

We extracted the collocations in the Brown Corpus section of the DSO corpus and, using 10-fold cross-validation, tagged the same corpus. Training and testing examples were thus from the same corpus. The same procedure was followed for the WSJ part. The results are shown in Tables 4 and 5. We can observe the following:

- The best kinds of collocations are **local content word** collocations, especially if two words from the context are taken into consideration, but the coverage is low. Function words to right and left also attain remarkable precision.
- Collocations are **stronger in the WSJ**, surely due to the fact that the BC is balanced, and therefore includes more genres and topics. This is a first indicator than **genre and topic** variations have to be taken into account.
- Collocations for fine-grained word-senses are **sensibly weaker** than those reported by Yarowsky (1993) for two-way ambiguous words. Yarowsky reports 99% precision, while our highest results do not reach 80%. It has to be noted that the test and training examples come from the same corpus, which means that, for some test cases, there are training examples from the same document. In some sense we can say that *one sense per discourse* comes into play. This point will be further explored in Section 7.

| Collocations | N Pr. | N Cov. | V Pr. | V Cov. | Overall Pr. | Overall Cov. |
|---|---|---|---|---|---|---|
| Word-to-right | .768 | .254 | .529 | .264 | .680 | .258 |
| Word-to-left | .724 | .185 | .867 | .182 | .775 | .184 |
| Two-words-to-right | .784 | .191 | .623 | .113 | .744 | .163 |
| Two-words-to-left | .811 | .160 | .862 | .179 | .830 | .166 |
| Word-to-right-and-left | .820 | .169 | .728 | .129 | .793 | .155 |
| Overall local content | .764 | .502 | .737 | .497 | .755 | .500 |
| Word-to-right | .600 | .457 | .527 | .370 | .577 | .426 |
| Word-to-left | .545 | .609 | .629 | .472 | .570 | .560 |
| Two-words-to-right | .638 | .133 | .687 | .084 | .650 | .116 |
| Two-words-to-left | .600 | .140 | .657 | .108 | .617 | .128 |
| Word-to-right-and-left | .721 | .220 | .694 | .138 | .714 | .191 |
| PoS-to-right | .490 | .993 | .488 | .993 | .489 | .993 |
| PoS-to-left | .465 | .991 | .584 | .994 | .508 | .992 |
| Two-PoS-to-right | .526 | .918 | .534 | .879 | .529 | .904 |
| Two-PoS-to-left | .518 | .822 | .614 | .912 | .555 | .854 |
| PoS-to-right-and-left | .555 | .918 | .634 | .891 | .583 | .908 |
| Overall local PoS&Fun | .622 | 1.00 | .640 | 1.00 | .629 | 1.00 |
| Word in sentence | .611 | 1.00 | .572 | 1.00 | .597 | 1.00 |
| Word in Window of 4 | .627 | .979 | .611 | .975 | .622 | .977 |
| Overall global content | .617 | 1.00 | .580 | 1.00 | .604 | 1.00 |
| OVERALL | .661 | 1.00 | .635 | 1.00 | .652 | 1.00 |

**Table 4:** Train on WSJ, tag WSJ.

| Collocations | N Pr. | N Cov. | V Pr. | V Cov. | Overall Pr. | Overall Cov. |
|---|---|---|---|---|---|---|
| Word-to-right | .644 | .203 | .432 | .230 | .562 | .212 |
| Word-to-left | .626 | .124 | .770 | .139 | .681 | .129 |
| Two-words-to-right | .657 | .146 | .500 | .103 | .613 | .131 |
| Two-words-to-left | .740 | .092 | .819 | .122 | .774 | .103 |
| Word-to-right-and-left | .647 | .088 | .686 | .114 | .663 | .098 |
| Overall local content | .675 | .405 | .635 | .404 | .661 | .405 |
| Word-to-right | .480 | .503 | .452 | .406 | .471 | .468 |
| Word-to-left | .414 | .639 | .572 | .527 | .464 | .599 |
| Two-words-to-right | .520 | .183 | .624 | .113 | .547 | .158 |
| Two-words-to-left | .420 | .131 | .648 | .173 | .516 | .146 |
| Word-to-right-and-left | .549 | .238 | .654 | .160 | .577 | .210 |
| PoS-to-right | .340 | .992 | .356 | .992 | .346 | .992 |
| PoS-to-left | .350 | .994 | .483 | .992 | .398 | .993 |
| Two-PoS-to-right | .406 | .923 | .422 | .876 | .412 | .906 |
| Two-PoS-to-left | .396 | .792 | .539 | .897 | .452 | .829 |
| PoS-to-right-and-left | .416 | .921 | .545 | .885 | .461 | .908 |
| Overall local PoS&Fun | .486 | 1.00 | .560 | 1.00 | .512 | 1.00 |
| Word in sentence | .545 | 1.00 | .492 | 1.00 | .526 | 1.00 |
| Word in Window of 4 | .550 | .972 | .525 | .951 | .541 | .964 |
| Overall global content | .549 | 1.00 | .503 | 1.00 | .533 | 1.00 |
| OVERALL | .577 | 1.00 | .564 | 1.00 | .572 | 1.00 |

**Table 5:** Train on BC, tag BC.

1. state -- (*the group of people comprising the government of a sovereign*)
2. state, province
   -- (*the territory occupied by one of the constituent administrative districts of a nation*)
3. state, nation, country, land, commonwealth, res publica, body politic
   -- (*a politically organized body of people under a single government*)
4. state -- (*the way something is with respect to its main attributes*)
5. Department of State, State Department, State
   -- (*the federal department that sets and maintains foreign policies*)
6. country, state, land, nation -- (*the territory occupied by a nation*)

**Figure 1:** Word senses for *state* in WordNet 1.6 (6 out of 8 are shown)

In the rest of this paper, only the overall results for each subset of the collocations will be shown. We will pay special attention to local-content collocations, as they are the strongest, and also closer to strict definitions of collocation.

As an example of the learned collocations Table 6 shows some strong local content word collocations for the noun *state*, and Figure 1 shows the word senses of *state* (6 out of the 8 senses are shown as the rest were not present in the corpora).

## 6 Cross-corpora experiments: one sense per collocation in doubt.

In these experiments we train on the Brown Corpus and tag the WSJ corpus and vice versa. Tables 7 and 8, when compared to Tables 4 and 5 show a significant drop in performance (both precision and coverage) for all kind of collocations (we only show the results for each subset of collocations). For instance, Table 7 shows a drop in .16 in precision for local content collocations when compared to Table 4.

These results confirm those by (Escudero et al. 2000) who conclude that the information learned in one corpus is not useful to tag the other.

In order to analyze the reason of this performance degradation, we compared the local content-word collocations extracted from one corpus and the other. Table 9 shows the amount of collocations extracted from each corpus, how many of the collocations are shared on average and how many of the shared collocations are in contradiction. The low amount of collocations shared between both corpora could explain the poor figures, but for some words (e.g. ***point***) there is a worrying proportion of contradicting collocations.

We inspected some of the contradicting collocations and saw that in all the cases they were caused by errors (or at least differing criteria) of the hand-taggers when dealing with words with difficult sense distinctions. For instance, Table 10 shows some collocations of *point* which receive contradictory senses in the BC and the WSJ. The collocation *important point*, for instance, is assigned the second sense[1] in all 3 occurrences in the BC, and the fourth sense[2] in all 2 occurrences in the WSJ.

| Collocations | Log | Senses | | | | | |
|---|---|---|---|---|---|---|---|
| | | #1 | #2 | #3 | #4 | #5 | #6 |
| State government | 3.68 | - | - | - | - | 4 | - |
| six states | 3.68 | - | - | - | - | 4 | - |
| State 's largest | 3.68 | - | - | - | - | 4 | - |
| State of emergency | 3.68 | - | 4 | - | - | - | - |
| Federal , state | 3.68 | - | - | - | - | 4 | - |
| State , including | 3.68 | - | - | - | - | 4 | - |
| Current state of | 3.40 | - | 3 | - | - | - | - |
| State aid | 3.40 | - | - | - | 3 | - | - |
| State where Farmers | 3.40 | 3 | - | - | - | - | - |
| State of mind | 3.40 | - | 3 | - | - | - | - |
| Current state | 3.40 | - | 3 | - | - | - | - |
| State thrift | 3.40 | - | - | - | 3 | - | - |
| Distributable state aid | 3.40 | - | - | - | 3 | - | - |
| State judges | 3.40 | - | - | - | - | 3 | - |
| a state court | 3.40 | - | - | 3 | - | - | - |
| said the state | 3.40 | - | - | - | - | 3 | - |
| Several states | 3.40 | - | - | - | - | 3 | - |
| State monopolies | 3.40 | - | - | - | 3 | - | - |
| State laws | 3.40 | - | - | 3 | - | - | - |
| State aid bonds | 3.40 | - | - | - | 3 | - | - |
| Distributable state | 3.40 | - | - | - | 3 | - | - |
| State and local | 2.01 | - | - | 1 | 1 | 15 | - |
| Federal and state | 1.60 | - | - | - | 1 | 5 | - |
| State court | 1.38 | - | - | 12 | - | 3 | - |
| Other state . | 1.38 | 4 | - | - | - | 1 | - |
| State governments | 1.09 | 1 | - | - | - | 3 | - |

**Table 6:** Local content-word collocations for *State* in WSJ

| | N | | V | | Overall | |
|---|---|---|---|---|---|---|
| Collocations | Pr. | Cov. | Pr. | Cov. | Pr. | Cov. |
| Overall local content | .597 | .338 | .591 | .356 | .595 | .344 |
| Overall local PoS&Fun | .478 | .999 | .491 | .997 | .483 | .998 |
| Overall global content | .442 | 1.00 | .455 | .999 | .447 | 1.00 |
| OVERALL | .485 | 1.00 | .497 | 1.00 | .489 | 1.00 |

**Table 7:** Train on BC, tag WSJ

| | N | | V | | Overall | |
|---|---|---|---|---|---|---|
| Collocations | Pr. | Cov. | Pr. | Cov. | Pr. | Cov. |
| Overall local content | .512 | .273 | .556 | .336 | .530 | .295 |
| Overall local PoS&Fun | .421 | 1.00 | .486 | 1.00 | .444 | 1.00 |
| Overall global content | .392 | 1.00 | .423 | 1.00 | .403 | 1.00 |
| OVERALL | .429 | 1.00 | .483 | 1.00 | .448 | 1.00 |

**Table 8:** Train on WSJ, tag BC

We can therefore conclude that the **one sense per collocation holds across corpora**, as the contradictions found were due to tagging errors. The low amount of collocations in common would explain in itself the low figures on cross-corpora tagging.

But yet, we wanted to further study the reasons of the **low number of collocations in common**, which causes the low cross-corpora performance. We thought of several factors that could come into play:

a) As noted earlier, the training and test examples from the in-corpus experiments are taken at random, and they could be drawn from the same document. This could make the results appear better for in-corpora experiments. On the contrary, in the cross-corpora experiments training and testing example come from different documents.
b) The genre and topic changes caused by the shift from one corpus to the other.
c) Corpora have intrinsic features that cannot be captured by sole genre and topic variations.
d) The size of the data, being small, would account for the low amount of collocations shared.

We explore a) in Section 7 and b) in Section 8.
c) and d) are commented in Section 8.

## 7 Drawing training and testing examples from the same documents affects performance

In order to test whether drawing training and testing examples from the same document or not explains the different performance in in-corpora and cross-corpora tagging, low cross-corpora results, we performed the following experiment. Instead of organizing the 10 random subsets for cross-validation on the *examples*, we choose 10 subsets of the *documents* (also at random). This way, the testing examples and training examples are guaranteed to come from different documents. We also think that this experiment would show more realistic performance figures, as a real application can not expect to find examples from the documents used for training.

---

[1] The second sense of *point* is defined as *the precise location of something; a spatially limited location*.
[2] Defined as *an isolated fact that is considered separately from the whole*.

| Word | PoS | # Coll. BC | # Coll. WSJ | % Coll. Shared | % Coll. Contradict. |
|---|---|---|---|---|---|
| Age | N | 45 | 60 | 27 | 0 |
| Art | N | 24 | 35 | 34 | 20 |
| Body | N | 12 | 20 | 12 | 0 |
| Car | N | 92 | 99 | 17 | 0 |
| Child | N | 77 | 111 | 40 | 05 |
| Cost | N | 88 | 88 | 32 | 0 |
| Head | N | 77 | 95 | 07 | 33 |
| Interest | N | 80 | 141 | 32 | 33 |
| Line | N | 110 | 145 | 20 | 38 |
| Point | N | 44 | 44 | 32 | 86 |
| State | N | 196 | 214 | 28 | 48 |
| Thing | N | 197 | 183 | 66 | 52 |
| Work | N | 112 | 149 | 46 | 63 |
| Become | V | 182 | 225 | 51 | 15 |
| Fall | V | 36 | 68 | 19 | 60 |
| Grow | V | 61 | 71 | 36 | 33 |
| Lose | V | 63 | 56 | 47 | 43 |
| Set | V | 94 | 113 | 54 | 43 |
| Speak | V | 34 | 38 | 28 | 0 |
| Strike | V | 12 | 17 | 14 | 0 |
| Tell | V | 137 | 190 | 45 | 57 |

**Table 9:** Collocations shared and in contradiction between BC and WSJ.

| Collocation | BC #2 | #4 | Other | WSJ #2 | #4 | Other |
|---|---|---|---|---|---|---|
| important point | 3 | 0 | 0 | 0 | 2 | 0 |
| point of view | 1 | 13 | 1 | 19 | 0 | 0 |

**Table 10:** Contradictory senses of *point*

Unfortunately, there are not any explicit document boundaries, neither in the BC nor in the WSJ.

In the BC, we took files as documents, even if files might contain more than one excerpt from different documents. This guarantees that document boundaries are not crossed. It has to be noted that following this organization, the target examples would share fewer examples from the same topic. The 168 files from the BC were divided in 10 subsets at random: we took 8 subsets with 17 files and 2 subsets with 16 files.

For the WSJ, the only cue was the directory organization. In this case we were unsure about the meaning of this organization, but hand inspection showed that document boundaries were not crossing discourse boundaries. The 61 directories were divided in 9 subsets with 6 directories and 1 subset with 7.

Again, 10-fold cross-validation was used, on these subsets and the results in Tables 11 and 12 were obtained. The Δ column shows the change in precision with respect to Tables 5 and 6.

Table 12 shows that, for the BC, precision and coverage, compared to Table 5, are degraded significantly. On the contrary results for the WSJ are nearly the same (cf. Tables 11 and 4).

The results for WSJ indicate that drawing training and testing data from the same or different documents in itself does not affect so much the results. On the other hand, the results for BC do degrade significantly. This could be explained by the greater variation in topic and genre between the files in the BC corpus. This will be further studied in Section 8.

Table 13 summarizes the overall results on WSJ and BC for each of the different experiments performed. The figures show that drawing training and testing data from the same or different documents would not in any case explain the low figures in cross-corpora tagging.

## 8 Genre and topic variation affects performance

Trying to shed some light on this issue we observed that the category *press:reportage*, is related to the genre/topics of the WSJ. We therefore designed the following experiment: we tagged each category in the BC with the decision lists trained on the WSJ, and also with the decision lists trained on the rest of the categories in the BC.

Table 14 shows that the local content-word collocations trained in the WSJ attain the best precision and coverage for *press:reportage*, both compared to the results for the other categories, and to the results attained by the rest of the BC on *press:reportage*. That is:

- From all the categories, the collocations from *press:reportage* are the most similar to those of WSJ.
- WSJ contains collocations which are closer to those of *press:reportage*, than those from the rest of the BC.

In other words, having related genre/topics help having common collocations, and therefore, warrant better word sense disambiguation performance.

|   | Overall | | | Local content | | |
|---|---|---|---|---|---|---|
|   | pr. | cov. | Δ pr. | pr. | cov. | Δ pr. |
| N | .650 | 1.00 | -.011 | .762 | .486 | -.002 |
| V | .634 | 1.00 | -.001 | .697 | .494 | -.040 |
| Overall | .644 | 1.00 | -.011 | .738 | .489 | -.017 |

**Table 11:** Train on WSJ, tag WSJ, crossvalidation according to files

|   | Overall | | | Local content | | |
|---|---|---|---|---|---|---|
|   | pr. | cov. | Δ pr. | pr. | cov. | Δ pr. |
| N | .499 | 1.00 | -.078 | .573 | .307 | -.102 |
| V | .543 | 1.00 | -.021 | .608 | .379 | -.027 |
| Overall | .514 | 1.00 | -.058 | .587 | .333 | -.074 |

**Table 12:** Train on BC, tag BC, crossvalidation according to files

|   | Overall (prec.) | | |
|---|---|---|---|
|   | In-corpora (examples) | In-corpora (files) | Cross-corpora |
| WSJ | .652 | .644 | .489 |
| BC | .572 | .514 | .448 |

**Table 13:** Overall results in different experiments

| Category | WSJ local content | | Rest of BC local content | |
|---|---|---|---|---|
|   | pr. | cov. | pr. | cov. |
| Press: Reportage | **.625** | **.330** | .541 | .285 |
| Press: Editorial | .504 | .283 | **.593** | .334 |
| Press: Reviews | .438 | .268 | **.488** | .404 |
| Religion | .409 | .306 | **.537** | .326 |
| Skills and Hobbies | .569 | .296 | **.571** | .302 |
| Popular Lore | .488 | .304 | **.563** | .353 |
| Belles Lettres, ... | .516 | .272 | **.524** | .314 |
| Miscellaneous | **.534** | .321 | **.534** | .304 |
| Learned | .518 | .257 | **.563** | .280 |
| General Fiction | .525 | .239 | **.605** | .321 |
| Mystery and ... | .523 | .243 | **.618** | .369 |
| Science Fiction | .459 | .211 | **.586** | .307 |
| Adventure and ... | .551 | .223 | **.702** | .312 |
| Romance and ... | .561 | .271 | **.595** | .340 |
| Humor | .516 | .321 | **.524** | .337 |

**Table 14:** Tagging different categories in BC. Best precision results are shown in bold.

## 9 Reasons for cross-corpora degradation

The goal of sections 7 and 8 was to explore the possible causes for the **low number of collocations in common** between BC and WSJ. Section 7 concludes that drawing the examples from different files is not the main reason for the degradation. This is specially true when the corpus has low genre/topic variation (e.g. WSJ). Section 8 shows that sharing genre/topic is a key factor, as the WSJ corpus attains better results on the *press:reportage* category than the rest of the categories on the BC itself. Texts on the same genre/topic share more collocations than texts on disparate genre/topics, even if they come from different corpora.

This seems to also rule out explanation c) (cf. Section 6), as a good measure of topic/genre similarity would help overcome cross-corpora problems.

That only leaves the low amount of data available for this study (explanation d). It is true that data-scarcity can affect the number of collocations shared across corpora. We think that larger amounts will make this number grow, especially if the corpus draws texts from different genres and topics. Nevertheless, the figures in Table 14 indicate that even in those conditions genre/topic relatedness would help to find common collocations.

## 10 Conclusions

This paper shows that the one sense per collocation hypothesis is weaker for fine-grained word sense distinctions (e.g. those in WordNet): from the 99% precision mentioned for 2-way ambiguities in (Yarowsky, 1993) we drop to 70% figures. These figures could perhaps be improved using more available data.

We also show that one sense per collocation does hold across corpora, but that collocations vary from one corpus to other, following genre and topic variations. This explains the low results when performing word sense disambiguation across corpora. In fact, we demonstrated that when two independent corpora share a related genre/topic, the word sense disambiguation results would be better.

This has considerable impact in future work on word sense disambiguation, as genre and topic are shown to be crucial parameters. A system trained on a specific genre/topic would have difficulties to adapt to new genre/topics. Besides, methods that try to extend automatically the amount of examples for training need also to account for genre and topic variations.

As a side effect, we have shown that the results on usual WSD exercises, which mix training and test data drawn from the same documents, are higher than those from a more realistic setting.

We also discovered several hand-tagging errors, which distorted extracted collocations. We did not evaluate the extent of these errors, but they certainly affected the performance on cross-corpora tagging.

Further work will focus on evaluating the separate weight of genre and topic in word sense disambiguation performance, and on studying the behavior of each particular word and features through genre and topic variations. We plan to devise ways to integrate genre/topic parameters into the word sense disambiguation models, and to apply them on a system to acquire training examples automatically.